%% file: root.tex
\documentclass[a4paper,conference]{IEEEtran}

\usepackage{multirow, graphicx}
\usepackage{amsfonts}
\usepackage{amsmath}
\usepackage{dsfont}
\usepackage{xcolor}

\begin{document}

\title{Penalizing Proposals using Classifiers for Semi-Supervised Object Detection}

\author{\IEEEauthorblockN{Somnath Hazra}
\IEEEauthorblockA{Indian Institute of Technology, Kharagpur\\
Kharagpur, India\\
Email: somnathhazra@kgpian.iitkgp.ac.in}
\and
\IEEEauthorblockN{Pallab Dasgupta}
\IEEEauthorblockA{Indian Institute of Technology, Kharagpur\\
Kharagpur, India\\
Email: pallab@cse.iitkgp.ac.in}}

\maketitle

\begin{abstract}
   Obtaining gold standard annotated data for object detection is often costly, involving human-level effort. Semi-supervised object detection algorithms solve the problem with a small amount of gold-standard labels and a large unlabelled dataset used to generate silver-standard labels. But training on the silver standard labels does not produce good results, because they are machine-generated annotations. In this work, we design a modified loss function to train on large silver standard annotated sets generated by a weak annotator. We include a confidence metric associated with the annotation as an additional term in the loss function, signifying the quality of the annotation. We test the effectiveness of our approach on various test sets and use numerous variations to compare the results with some of the current approaches to object detection. In comparison with the baseline where no confidence metric is used, we achieved a 4\% gain in mAP with 25\% labeled data and 10\% gain in mAP with 50\% labeled data by using the proposed confidence metric. \footnote{Under consideration at Computer Vision and Image Understanding}
\end{abstract}

\input{sections/introduction}
\input{sections/literature}
\input{sections/background}
\input{sections/approximate}
\input{sections/experiments}
\input{sections/results}
\input{sections/conclusion}

\bibliographystyle{IEEEtran}
\bibliography{ref}

\end{document}

%% file: sections/introduction.tex
\section{Introduction}
\label{sec1}

Object detection belongs to a class of computer vision algorithms that help locate objects within an image by drawing bounding boxes around them. Deep Learning algorithms, specifically Convolutional Neural Networks (CNNs) have had a significant impact in the field in the recent past due to their ability to produce good accuracy owing to their generalization capability. But, in general, the algorithms require adequate training data to achieve good performance. The situation stands similar for object detection algorithms also, like YOLO (You Only Look Once) \cite{redmon2016you} and R-CNN (Region Convolution Neural Network) \cite{girshick2014rich}. Accumulating gold standard annotated data takes substantial human effort for object detection algorithms. Although there have been recent improvements regarding self-supervised learning in this field \cite{lee2019multi, liu2020self, amrani2020self}; the approaches either do not perform well or are specific to the area of application.

Semi-Supervised Learning (SSL) is a learning paradigm requiring a small number of labeled samples and a large number of unlabeled samples. It is useful in situations where gold standard data is difficult to obtain and involves human-level supervision \cite{su2012crowdsourcing}; as is the problem in object detection. Current approaches to semi-supervised learning for object detection rely on proposal labels generated by an annotator (also called teacher) trained on a small number of gold standard annotations \cite{sohn2020simple, xu2021end, adhikari2021iterative}. The generated annotations however may contain errors and inconsistencies. These problems are solved either by using filters to separate and include the more accurate labels into the silver standard set; or by calling for manual intervention for the correction of anomalies. Using filters to separate trainable \emph{teacher annotations} may discard a lot of generated labels. Manual corrections however reduce the number of discarded samples but include manual effort in the process, which we aim to avoid altogether.

\begin{figure}[!t]
    \centering
    \includegraphics[width=2.7in]{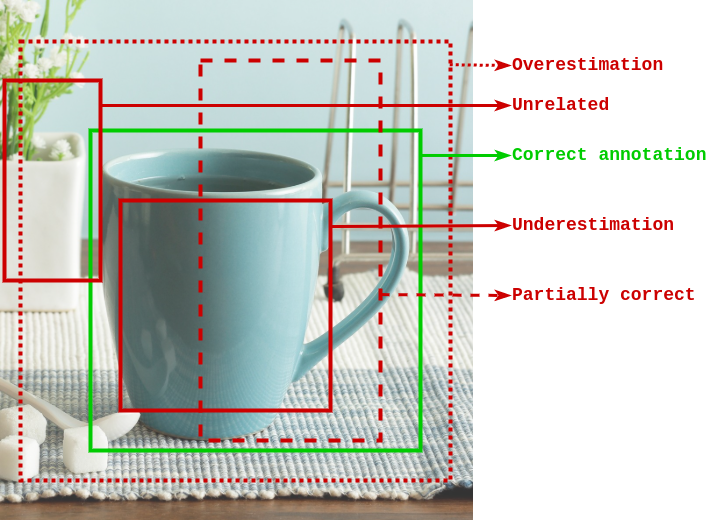}
    \caption{Sample image showing some of the annotation errors in the labels generated by the teacher network.}
    \label{fig1}
\end{figure}

To solve the above problems, firstly, we use all the silver standard annotations generated by the teacher, without discarding the imperfect ones. The blank annotations, however, have to be removed. It is assumed that we have abundant unlabeled samples to generate a large enough silver standard dataset, and the problem at hand is to generate labels distinctly for the task of object detection only. Secondly, as demonstrated in Figure \ref{fig1}, the generated annotations contain many types of anomalies that may not result in a good performance. Thus for using all the samples we need to use some feedback mechanism from classifiers that assigns a score to the labels, which gives an intuition as to how good the labels are. To this end, we propose various scoring mechanisms which do not require manual intervention in the loop, reason their usage and perform various ablation tests to see what performs better. We define the method to include the feedback scores in the learning process. Our algorithm is simple enough to be integrated with any learning paradigm.

As an initial step of the process, we train an object detection algorithm with the gold standard labels; which form the teacher. The trained detector is used to generate pseudo labels or teacher annotations from a large unlabelled set post-application of non-max suppression (NMS). The pseudo labels are then scored using a pre-trained classifier. The process is followed by the final training and evaluation process using the pseudo labels along with their corresponding scores obtained from the classifiers.

We present the efficiency of our algorithm using the backbone architecture of the YOLOv3 (40M params), EfficientDet (8M params) and YOLOX (9M params) object detection frameworks \cite{redmon2018yolov3, tan2020efficientdet, ge2021yolox}. We obtained the gold standard training set and test set from the Open Images and Pascal VOC datasets \cite{kuznetsova2020open, pascal-voc-2012} and our unlabelled samples from ImageNet \cite{deng2009imagenet} and some other open sources. We show the effect of reduction in the size of the gold standard training data on the number of generated annotations and also on the overall performance of the algorithm.

The contributions of this text can be summarized with the following points:

\begin{enumerate}
    \item We propose a feedback mechanism from classifiers that automatically assigns a score to each of the silver standard labels, signifying the quality of the label.
    \item The quality label is included in the training process through the loss function with the help of a hyperparameter, which requires no additional tuning and is simple enough to be extended to other backbone architectures. We demonstrated using YOLOv3, EfficientDet and YOLOX as our backbone.
    \item We tested the efficiency of our algorithm by comparing the results obtained on some of the current algorithms in the field of object detection, using the Open Images, VOC2012 datasets for the gold-standard annotations. The algorithm achieved up to 10\% gain in mAP (mean Average Precision) over the baseline as discussed in Section \ref{sec6}.
\end{enumerate}

%% file: sections/literature.tex
\section{Related Works}
\label{sec2}

\subsection{Object Detection}

Object detection algorithms in deep learning literature can be primarily classified under two threads: (i) Single-stage methods, (ii) Two-stage methods. Most of the two-stage methods \cite{girshick2014rich, girshick2015fast, ren2015faster, lin2017feature, he2017mask} are based on defining regions of interest that separate the probable object pixels using segmentation algorithms or CNNs. This is followed by the classification of the proposed regions and bounding box generation. These algorithms are slower than the single-stage methods and are generally difficult to optimize \cite{redmon2016you}. Single-stage algorithms like YOLO and its variants, SSD, etc. \cite{redmon2016you, liu2016ssd,redmon2017yolo9000, redmon2018yolov3, tan2020efficientdet} use CNNs with a regression head for generating bounding boxes and a classification head for determining the class of object present in the box. The two heads operate in parallel which makes the algorithms comparatively faster, with variants such as \cite{redmon2017yolo9000} used for real-time detection. But all the algorithms require a large amount of annotated data for giving good results, which we try to solve with semi-supervised methods.

\subsection{Semi-Supervised Object Detection}

Semi-supervised learning literature for images describes two kinds of methodologies. Consistency regularization methods \cite{sajjadi2016mutual, tarvainen2017mean, berthelot2019mixmatch, jeong2019consistency, gao2020consistency} are based on preserving consistency in predictions between models with perturbed inputs that do not change the label. There are adaptations of the algorithm in the object detection domain \cite{jeong2019consistency} that train the detector on two loss functions depending on whether the sample was previously labeled. Self-training methods \cite{lee2013pseudo, li2019learning, xie2020self, sohn2020simple} are another class of algorithms using pseudo labels generated by a soft teacher to train on a large number of unlabeled samples. There are many studies involving the generation of pseudo labels in the object detection literature \cite{sohn2020simple, xu2021end, chauhan2021semi, yang2021interactive}. Some of these algorithms involve filtration of the pseudo labels based on some threshold \cite{sohn2020simple, xu2021end}, to discard the labels that may not be useful. We eliminate the need for the removal of a lot of pseudo labels and propose a scoring mechanism indicating the quality of the generated labels. Most of the algorithms are based on the R-CNN backbone, which is comparatively slower and less accurate than single-stage algorithms like YOLO. Although we explain the algorithm and show the results obtained using the YOLOv3, YOLOX backbones, our algorithm is simple enough that can be extended to many other architectures.

Some texts suggest human supervision for correction of teacher annotations post the first phase of training \cite{adhikari2018faster, adhikari2021iterative, adhikari2021sample} as in the previous texts. The teacher algorithm is trained on the gold standard samples and generates the pseudo labels for the unlabeled set in the form of batches. The labels are corrected with the help of manual inspection and correction and used to train the algorithm in the second phase. In these texts, the pseudo labels are thus passed through a quality control filter which is performed under human supervision. We aim to achieve a similar objective automatically by integrating the quality scores with the second phase of the learning process; thus avoiding the need for manual intervention altogether.

%% file: sections/background.tex
\section{Background}
\label{sec3}

In this section, we briefly explain the algorithm and the loss function of general backbone detectors, using the example of YOLO. We also introduce the method to integrate the quality score in the training.


YOLO \cite{redmon2016you} belongs to a family of single-shot object detection algorithms that use convolutions to parallelly predict bounding boxes and class confidence scores using two heads. The algorithm divides the input image into a system of $S \times S$ grid. For each grid cell; the classification head predicts $C$ class probabilities conditioned on the presence of an object; and the regression head predicts $B$ bounding boxes for each grid cell, i.e., their position, width, and height, along with a confidence score. We used YOLOv3, a popular industry standard in the field; and YOLOX, the most recent addition.

\subsection{The Loss Function}

The loss function for YOLO consists of five parts based on multiple optimization objectives. For a single training sample, the loss function is given by \cite{redmon2016you}:

\begin{align*}
    & \; \lambda_{coord} \sum_{i=0}^{S^2} \sum_{j=0}^B \mathds{1}_{ij}^{obj} [(x_i - \hat{x}_i)^2 + (y_i - \hat{y}_i)^2] && : \textit{Centre}\\
    + & \; \lambda_{coord} \sum_{i=0}^{S^2} \sum_{j=0}^B \mathds{1}_{ij}^{obj} [(w_i^{.5} - \hat{w}_i^{.5})^2 + (h_i^{.5} - \hat{h}_i^{.5})^2] && : \textit{Box}\\
    + & \; \sum_{i=0}^{S^2} \sum_{j=0}^B \mathds{1}_{ij}^{obj} (C_i - \hat{C}_i)^2 && : \textit{Object}\\
    + & \; \lambda_{noobj} \sum_{i=0}^{S^2} \sum_{j=0}^B \mathds{1}_{ij}^{noobj} (C_i - \hat{C}_i)^2 && : \textit{No obj}\\
    + & \; \sum_{i=0}^{S^2} \mathds{1}_{i}^{obj} \sum_{c \in classes} (p_i(c) - \hat{p}_i(c))^2 && : \textit{Score}
\end{align*}

where $\mathds{1}_i^{obj}$ denotes presence of object in cell $i$ and $\mathds{1}_{ij}^{obj}$ denotes the bounding box predictor $j$ in cell $i$ that is responsible for that prediction. The variables $(x, y), w, h$ correspond to the centre co-ordinates and dimensions of the bounding box.

\subsection{Integrating the confidence metric}

The objective function does not consider the quality of the annotated data used for training; i.e., it does not contemplate how well the annotated bounding box encloses the object of interest. This is because it is assumed that the core annotations are done by humans and are what we consider to be gold standard data. In our case, however, the dataset is heuristically generated or machine-generated. Pre-trained classifiers are therefore used here to appropriately balance the loss by introducing a metric to judge the quality of annotation. The details are provided in the following section.

%% file: sections/approximate.tex
\section{Approximating the Loss}
\label{sec4}

The silver standard teacher annotations that are generated may not be reliable enough as compared to human annotations. Penalizing the optimization objective based on these annotations leads to a situation of \emph{high variance}, limiting the generalization capacity of the detection algorithm. Therefore, an additional confidence score is incorporated to assess the quality of the silver standard annotation and accordingly penalize the objective function. Through this, we attempt to reduce the penalization by a factor that is indicative of the quality of silver standard annotation.

The loss function  described in Section \ref{sec3} may be represented by the following function:

\begin{equation}
    \label{eqn1}
    \mathcal{L}(\theta) = \sum_{i=0}^N \sum_{j=0}^M L(\hat{o}_{ij}, o_{ij} | \theta, I)
\end{equation}

where $N$ is the total number of annotated bounding boxes, and $M$ is the total number of losses corresponding to each box ($M = 5$ for YOLO). $L$ signifies the loss function. The term $\hat{o}_{ij}$ is a generic representation of any of the outputs from the network corresponding to bounding box $i$ and object $j$, the network parameters are represented by $\theta$, for an image $I$; $o_{ij}$ is the ground truth label. Equation (\ref{eqn1}) is an attempt to generalize the representation for the optimization objective used in any base architecture.

Now, the confidence metric is integrated into this loss function to directly optimize over a single criterion. Intuitively, this metric helps to marginalize updates in the direction of imperfectly annotated data; making the model robust to the obvious irregularities in the training data.

\subsection{Loss Function update}

The box predictions from the model in the first phase, given by $\bar{o}$ are used as ground truth labels for the second phase. We use the scores obtained from pre-trained classifiers as feedback on the quality of the label, i.e., how well the annotated box encloses the object as shown in Figure \ref{fig1}. A detailed explanation is provided below. The objective is to obtain better scores through the training process:

\begin{equation}
    \label{eqn2}
    \mathcal{L}(\theta) = - \mathbb{E}_{\bar{o} \sim f(\theta)} [\alpha(\bar{o})]
\end{equation}

Here, $\alpha(\bar{o})$ represents the classification score for the sampled boxes. Since the scoring function, $\alpha(\cdot)$ on the samples generated by the teacher, $\bar{o}$, is non-differentiable, the expected gradient of the function is estimated using the following equation:

\begin{equation}
    \label{eqn3}
    \nabla \mathcal{L}(\theta) = \mathbb{E}_{\bar{o} \sim f(\theta)} [\alpha(\bar{o}) \nabla L(\hat{o}, \bar{o} | \theta, I)]
\end{equation}

Equation (\ref{eqn1}) learns to maximize the likelihood of ground truth labels for the bounding boxes, while Equation (\ref{eqn2}) learns to select boxes that enclose the object better. The expected gradient in Equation (\ref{eqn3}) can be approximated by using the predicted bounding boxes for each training sample:

\begin{equation}
    \label{eqn4}
    \nabla \mathcal{L}(\theta) \approx \sum_{i=0}^N \alpha(\bar{o}_i) \sum_{j=0}^M \nabla L(\hat{o}_{ij}, \bar{o}_{ij} | \theta, I)
\end{equation}


We use non-max supression and confidence threshold to limit the output of the primary detection algorithm and output the more accurate samples.

\begin{figure*}[!ht]
    \centering
    \includegraphics[width=0.7\textwidth]{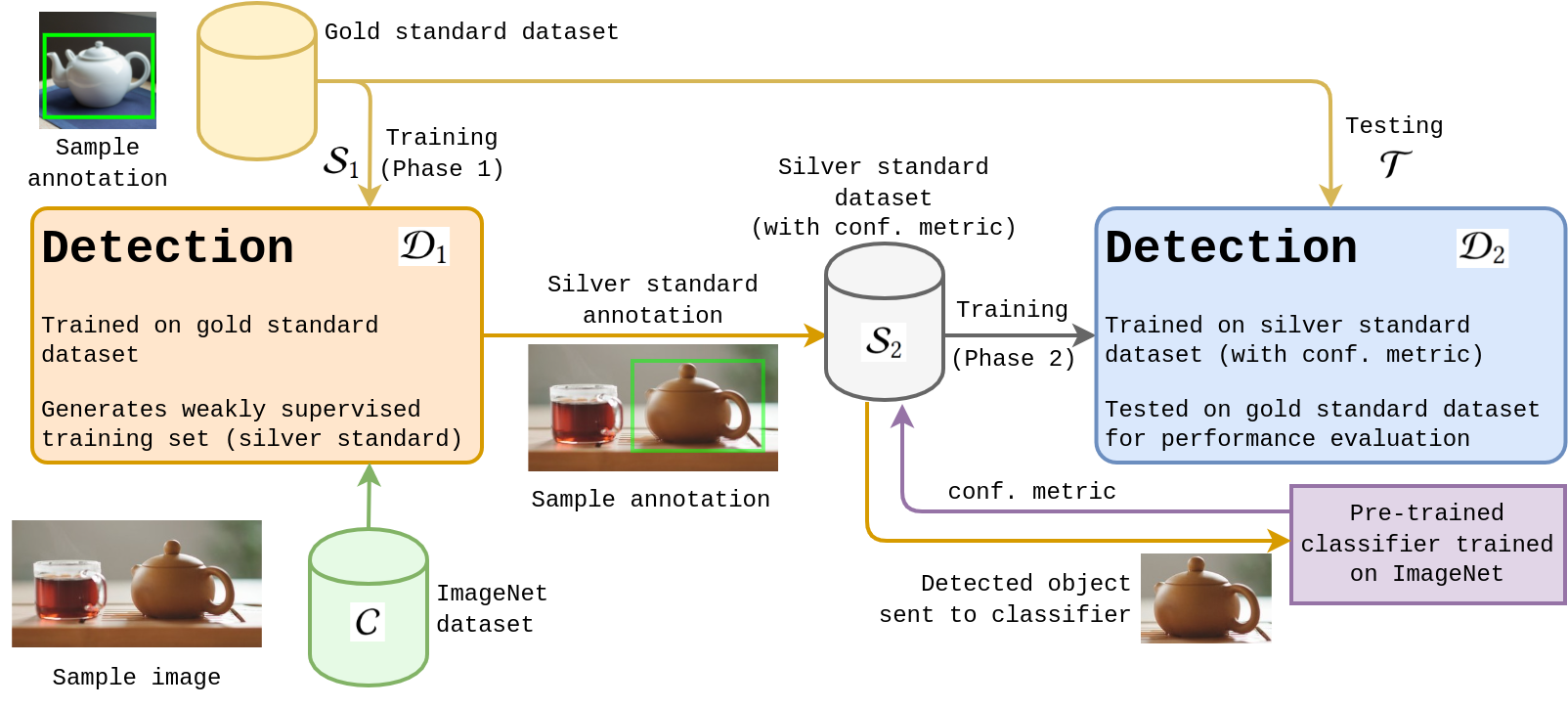}
    \caption{Overview of the pipeline. The dataset $\mathcal{S}_1$ was used to train $\mathcal{D}_1$; which generated detection labels for the images in $\mathcal{C}$, thus creating $\mathcal{S}_2$. The scores come from pre-trained classifiers, which validate the quality of the generated label by producing the classification score for the generated box label and are included in $\mathcal{S}_2$. The dataset $\mathcal{S}_2$ is used to train $\mathcal{D}_2$. Performance of $\mathcal{D}_2$ is tested using $\mathcal{T}$.}
    \label{fig2}
\end{figure*}

\subsection{Overall Methodology}

The overall algorithm is outlined in figure \ref{fig2}. The process involves the creation of a weakly supervised training set (silver standard) from a large set of images, $\mathcal{C}$, with the help of a detection algorithm $\mathcal{D}_1$ which is the teacher. The algorithm $\mathcal{D}_1$ is pre-trained on a comparatively small gold-standard object detection dataset (compared to silver-standard). We call the gold-standard dataset $\mathcal{S}_1$, and the silver standard dataset $\mathcal{S}_2$. The dataset $\mathcal{S}_2$ is generated by the teacher and contains what we call pseudo labels or teacher annotations. The goodness of the annotations generated by $\mathcal{D}_1$ is indicated by obtaining a classification score from the pre-trained classifiers on the generated bounding boxes for the corresponding label. We refer to this score as the \emph{confidence metric}. The scores are included in the dataset $\mathcal{S}_2$ which contains the images from the dataset $\mathcal{C}$ only.

The silver standard dataset $\mathcal{S}_2$ along with the corresponding scores are used for training the second phase of the detection algorithm $\mathcal{D}_2$. The performance of algorithm $\mathcal{D}_2$ with the new loss function is tracked using a part of the gold standard annotated data as test set, $\mathcal{T}$ that is disjoint from $\mathcal{S}_1$.

The intuition behind using the classifiers as a goodness assessor for the silver standard annotations is the classifiers' ability to score based on how well the bounding box covers the object of concern. That is, if the bounding box does not fully enclose the object, the annotation loses score since essential object pixels are lost. On the other hand, too large a box may cover other objects (e.g. the cup in Figure \ref{fig1}) resulting in a shift in score to other objects, owing to the use of Softmax activation in classifiers. If no other object is in the image, a larger bounding box results in loss of information from the object pixels, leading to lower scores. This is an issue that is tackled in the other texts involving attention networks \cite{ji2020attention, zhuang2020learning}, with the advent of tasks viz fine-grained visual categorization, specifically studied to challenge this problem; since ILSVRC networks are reported to perform poorly for these problems \cite{zhang2021multi}. Although not the best method to determine the quality of the labels and generates anomalies, as seen in Section \ref{sec6}; it automates the process, removing the manual effort required to surveil the generated samples.

%% file: sections/experiments.tex
\section{Experiments}
\label{sec5}

\subsection{Experimental Setup}

The experiments were conducted on two gold standard datasets separately: Open Images and VOC2012. The images from the Open Images Dataset was collected using OID toolkit \cite{OIDv4_ToolKit}. The images for set $\mathcal{C}$ that was used for the silver standard $\mathcal{S}_2$, was collected from ImageNet-1K dataset \cite{deng2009imagenet} and some other open sources. Each class consisted of $\sim$1K images on average. Images with successful detection by $\mathcal{D}_1$ were included in $\mathcal{S}_2$ post NMS. During label generation for the second phase of training, excessive imbalance in the training sets was removed to avoid any bias. We used data from five different classes for our experiments.

In the first experiment, the training set size of $\mathcal{S}_1$ was gradually reduced to note the variance in performance. This helped to identify the point of saturation in performance. The obtained scores are detailed in the following section.

The confidence scores for the teacher annotations, for both the experiments, were pre-computed and saved in $\mathcal{S}_2$. The classification scores for the bounded objects were obtained from 4 classifiers: AlexNet, VGG16, ResNet, and GoogLeNet, with $\sim$1M retraining parameters; all of which were pre-trained on the ImageNet dataset. Various ablation studies were conducted to empirically obtain the best scoring mechanism; as detailed in the following section.

\begin{table*}[!t]
\caption{Comparison of AP and mAP scores for multi-class object detection with variations in the size of $\mathcal{S}_1$. Here average and maximum of the scores for pre-trained classifiers were used as confidence metrics, with weights imported from gold standard pre-training which is explained in the ablation studies. Scores are reported for YOLOv3 \cite{redmon2018yolov3}, EfficientDet \cite{tan2020efficientdet} and YOLOX \cite{ge2021yolox}.}
\label{tab1}
\centering
\begin{tabular}{|l|l|l|l||c|c|c|c|c||c|}
\hline
\multirow{2}{*}{\textbf{Backbone}} &
\multirow{2}{*}{\textbf{Approach}} &
\multirow{2}{*}{\textbf{Dataset}} &
\multirow{2}{*}{\textbf{$\alpha(\bar{o}_i)$}} & \multicolumn{5}{c||}{\textbf{AP scores (\%)}}                                                 & \multirow{2}{*}{\textbf{mAP (\%)}} \\ \cline{5-9}
                                 &  &    &                                   & \textbf{teapot} & \textbf{monitor} & \textbf{mug} & \textbf{candle} & \textbf{icecream} &                               \\ \hline \hline
\multicolumn{10}{|l|}{\#samples per class in $\mathcal{S}_1$ = 256; average \#samples per class in $\mathcal{S}_2$ = 833}                                                                                                                                                         \\ \hline
\multirow{4}{*}{YOLOv3} & without feedback & gold standard & -                                     & 39.47           & 51.19            & 38.91        & 20.86           & 20.13             & 34.12                         \\ \cline{2-10}
& without feedback & silver standard & 1.0                                     & 47.55           & 50.63            & 41.72        & 18.06           & 27.76             & 37.14             \\ \cline{2-10}
& \multirow{2}{*}{with feedback} & \multirow{2}{*}{silver standard} & avg                                   & 53.47           & 54.53            & 41.53        & 16.54           & 25.54             & 38.32                         \\ \cline{4-10} 
                            &  &  & max                                   & 52.82           & 54.81            & 44.25        & 16.35           & 26.41             & 38.93                         \\ \hline
EfficientDet & without feedback & gold standard & -                                     & 58.20           & 52.00            & 56.40        & 16.90           & 21.80             & 41.10                      \\ \hline
\multirow{3}{*}{YOLOX} & without feedback  & gold standard & -                                     & 55.10           & 54.50            & 60.70        & 16.90           & 23.50             & 42.10                      \\ \cline{2-10}
& without feedback & silver standard & 1.0                                     & 63.50           & 58.20            & 67.00        & 21.00           & 28.60             & 47.60         \\ \cline{2-10}
& with feedback & silver standard & max                                     & 63.30           & 58.40            & 68.90        & 22.50           & 31.80             & \textbf{49.00}                      \\ \hline \hline
\multicolumn{10}{|l|}{\#samples per class in $\mathcal{S}_1$ = 224; average \#samples per class in $\mathcal{S}_2$ = 760}                                                                                                                                                         \\ \hline
\multirow{4}{*}{YOLOv3} & without feedback  & gold standard  & -                                     & 39.48           & 50.85            & 34.64        & 21.84           & 20.27             & 33.42                         \\ \cline{2-10}
& without feedback  & silver standard & 1.0                                     & 47.46           & 52.76            & 38.18        & 18.83           & 25.00             & 36.45                \\ \cline{2-10}
& \multirow{2}{*}{with feedback} & \multirow{2}{*}{silver standard} & avg                                   & 53.53           & 52.79            & 40.02        & 18.84           & 26.01             & 38.24                         \\ \cline{4-10} 
                             &   &  & max                                   & 51.50           & 52.57            & 42.84        & 16.36           & 24.57             & 37.57                         \\ \hline
EfficientDet & without feedback & gold standard & -                                     & 56.10           & 48.90            & 58.00        & 15.30           & 21.00             & 39.80                      \\ \hline
\multirow{3}{*}{YOLOX} & without feedback  & gold standard & -                                     & 53.90           & 51.70            & 59.70        & 15.90           & 23.20             & 40.90                      \\ \cline{2-10}
& without feedback & silver standard & 1.0                                     & 61.70           & 56.00            & 66.30        & 21.10           & 28.90             & 46.80               \\ \cline{2-10}
& with feedback & silver standard & max                                     & 64.40           & 57.40            & 66.30        & 21.30           & 30.10             & \textbf{47.90}                      \\ \hline \hline
\multicolumn{10}{|l|}{\#samples per class in $\mathcal{S}_1$ = 192; average \#samples per class in $\mathcal{S}_2$ = 676.4}                                                                                                                                                         \\ \hline
\multirow{4}{*}{YOLOv3} & without feedback  & gold standard & -                                     & 34.55           & 49.96            & 32.52        & 21.29           & 16.07             & 30.88                         \\ \cline{2-10}
& without feedback  & silver standard & 1.0                                     & 47.52           & 52.31            & 39.06        & 18.76           & 23.83             & 36.30             \\ \cline{2-10}
& \multirow{2}{*}{with feedback} & \multirow{2}{*}{silver standard} & avg                                   & 49.33           & 53.90            & 36.85        & 16.10           & 22.93             & 35.82                         \\ \cline{4-10} 
                        &     &  & max                                   & 50.37           & 52.57            & 39.98        & 15.96           & 26.79             & 37.14                         \\ \hline
EfficientDet & without feedback & gold standard & -                                     & 49.30           & 48.20            & 53.80        & 14.80           & 15.90             & 36.40                      \\ \hline
\multirow{3}{*}{YOLOX} & without feedback & gold standard & -                                     & 46.50           & 48.70            & 53.80        & 18.10           & 20.50             & 37.50                      \\ \cline{2-10}
& without feedback & silver standard & 1.0                                     & 61.30           & 56.00            & 63.70        & 19.90           & 28.10             & 45.80              \\ \cline{2-10}
& with feedback & silver standard & max                                     & 62.80           & 55.60            & 67.80        & 20.50           & 27.80             & \textbf{46.90}                      \\ \hline \hline
\multicolumn{10}{|l|}{\#samples per class in $\mathcal{S}_1$ = 160; average \#samples per class in $\mathcal{S}_2$ = 627.2}                                                                                                                                                         \\ \hline
\multirow{4}{*}{YOLOv3} & without feedback & gold standard & -                                     & 30.40           & 45.09            & 28.77        & 14.33           & 15.31             & 26.78                         \\ \cline{2-10}
& without feedback & silver standard & 1.0                                     & 44.88           & 54.03            & 28.99        & 15.28           & 23.96             & 33.43         \\ \cline{2-10}
& \multirow{2}{*}{with feedback} & \multirow{2}{*}{silver standard} & avg                                   & 44.35           & 53.73            & 31.37        & 15.41           & 27.77             & 34.53                         \\ \cline{4-10} 
                              &  &  & max                                   & 45.75           & 55.43            & 30.55        & 17.00           & 25.46             & 34.83                         \\ \hline
EfficientDet & without feedback & gold standard & -                                     & 43.80           & 43.60            & 48.50        & 14.80           & 15.10             & 33.20                      \\ \hline
\multirow{3}{*}{YOLOX} & without feedback & gold standard & -                                     & 51.90           & 44.50            & 41.50        & 13.80           & 15.90             & 33.50                      \\ \cline{2-10}
& without feedback & silver standard & 1.0                                     & 58.50           & 53.00            & 63.00        & 17.20           & 25.30             & 43.40                      \\ \cline{2-10}
& with feedback & silver standard & max                                     & 61.60           & 55.30            & 65.10        & 20.00           & 24.00             & \textbf{45.20}                      \\ \hline
\end{tabular}
\end{table*}

\subsection{Baseline Model}

As a baseline model for comparison, the same pipeline was used with $\alpha(\bar{o}_i)$ set to 1, i.e., the feedback is kept constant and independent of the quality of samples generated, as is the approach taken in some texts \cite{sohn2020simple, xu2021end, yang2021interactive} for self-training. The performances are also reported with training on the gold standard data, also with $\alpha(\bar{o}_i)$ set to 1. We also report the scores using the STAC algorithm \cite{sohn2020simple} on the VOC2012 dataset. The second phase of the training was carried out using the silver-standard dataset, $\mathcal{S}_2$, to indicate the improvement in performance compared to the gold-standard data.

The mAP scores are used as a comparative measure. It compares the precision of ground truth bounding boxes with the generated bounding boxes. It is based on Average Precision; which is calculated using IOU (Intersection over Union) for object detection algorithms.

%% file: sections/results.tex
\section{Results}
\label{sec6}

The results in Table \ref{tab1} show that for both YOLOX and YOLOv3, in most of the experiments better mAP scores were obtained when the classification score was used as the value of $\alpha(\bar{o}_i)$ as compared to considering $\alpha(\bar{o}_i) = 1$ or using the gold standard data. YOLOX generated the best scores with feedback, for all instances. The classifiers achieved an average accuracy of 92.47\% on the unlabelled dataset used for this experiment. For most cases, the `max' operation gave better scores than the `avg' operation \footnote{'max' = max(classification scores); 'avg' = avg(classification scores) from the four classifiers referred to in Setion \ref{sec5}. Detailed in Ablation Study.}. There were some exceptions, e.g., when $\mathcal{S}_1 = 192$ and $\alpha(\bar{o}_i) = \text{avg(classification scores)}$; and when $\mathcal{S}_1 = 224$. Also, for some classes, this observation deviates, common for the `candle' class. These anomalies may result from erroneous feedback from the classifiers or the varied differences between images in $\mathcal{S}_1$ and $\mathcal{C}$. For $\mathcal{S}_1 < 160$, the number of samples generated in $\mathcal{S}_2$ was very less, therefore not reported here. A larger unlabelled dataset may be useful in such circumstances.

\begin{table}[!t]
\centering

\caption{The mAP scores; $\mathcal{D}_1$ was trained on VOC2012. Scores of STAC on $\mathcal{S}_2$ are also reported for comparison. YOLOX used as the backbone gave the best results.}
\label{tab4}
\begin{tabular}{|l|l|l|c|}
\hline
\textbf{Algorithm}      & \textbf{Dataset} & $\alpha(\bar{o}_i)$ & {\textbf{mAP (\%)}} \\ \hline \hline
EfficientDet            & gold std.        & -    & 44.5                  \\ \hline
YOLOX                   & gold std.        & -    & 45.9                  \\ \hline
STAC                    & silver std.      & 2.0  & 49.7                 \\ \hline
with EfficientDet & silver std.      & 1.0  & 52.4                                       \\ \hline
Our (with EfficientDet) & silver std.      & max. & 55.1                                       \\ \hline
with YOLOX        & silver std.      & 1.0  & 53.4                                       \\ \hline
Out (with YOLOX)        & silver std.      & avg. & 55.4                                       \\ \hline
Out (with YOLOX)        & silver std.      & max. & \textbf{57.3}                                       \\ \hline
\end{tabular}

\end{table}

Comparing the performance of YOLOv3 using generated data to that of YOLOX and EfficientDet with gold-standard data, YOLOv3 gives better results in the bottom part of the table. With increased samples, the scores generated are more vicinal, compared to the scores generated by YOLOv3 using gold standard data only. The aim of this study was to show the performance of the algorithm on different objects and its effect overall. In the next section we empirically compare the performance of `max' and `avg' with other scores.

The second section of the experiments was performed with gold standard data obtained from VOC2012 \cite{pascal-voc-2012} on 5 classes ($\sim$1.5k gold labels). We compare the results with the STAC \cite{sohn2020simple} with Faster-RCNN as an additional baseline, to evaluate the performance. From the results in Table \ref{tab4}, the highest mAP is obtained using YOLOX\cite{ge2021yolox} as the backbone architecture. The classifiers achieved an average accuracy of 94.01\% on the classification labels, post retraining.

\subsection{Ablation Study}

\begin{table}[!t]
\caption{The mAP scores; $\mathcal{D}_1$ was trained on a dataset with 256 single class images. First the scores where the weights were reinitialised after completion of training of $\mathcal{D}_1$, then the weights were bootstrapped from the initial phase.}
\label{tab2}
\centering
\begin{tabular}{|l|l|c|}
\hline
\textbf{Approach} & \textbf{metric source} & \textbf{mAP (\%)} \\ \hline \hline
\multicolumn{3}{|l|}{Reinitialised weights after initial training}                        \\ \hline
No feedback               & -                                               & 53.15        \\ \hline
\multirow{5}{*}{With feedback} & YOLO classification score & 56.24        \\ \cline{2-3} 
                         & VGG16 classification score & 50.43        \\ \cline{2-3} 
                         & avg(classifiers' score) & 55.59        \\ \cline{2-3} 
                         & max(classifiers' score) & \textbf{57.52}        \\ \cline{2-3} 
                         & max(max, YOLO score) & 56.42        \\ \hline \hline
\multicolumn{3}{|l|}{Continued with weights from initial training}                        \\ \hline
No feedback                & -                                               & 53.69        \\ \hline
\multirow{5}{*}{With feedback} & YOLO classification score & 55.64        \\ \cline{2-3} 
                         & VGG16 classification score & 55.31        \\ \cline{2-3} 
                         & avg(classifiers' score) & 58.29        \\ \cline{2-3} 
                         & max(classifiers' score) & \textbf{59.24}        \\ \cline{2-3} 
                         & max(max, YOLO score) & 58.23        \\ \hline
\end{tabular}
\end{table}

We consider different classification scores for the value of $\alpha(\bar{o}_i)$ corresponding to an output $\bar{o}_i$ generated in the teacher annotation. The studies are limited to single class operations for the experiments in this section. The results in Table \ref{tab2} show that the `avg' and `max' of the classification scores from the four classifiers, when used as $\alpha(\bar{o}_i)$ produce the best results. We thus used it as a feedback mechanism for the multi-class operations. Better scores were obtained when the weights were bootstrapped from the initial phase of training. The gold standard score for this dataset was 55.92, which is better than both the scores when $\alpha(\bar{o}_i) = 1$. The phenomenon was observed in Table \ref{tab1} for our choice of values for $\alpha(\bar{o}_i)$ also. This shows that having more unlabeled data may not help if the feedback is not a proper reflection of the quality of annotation. The scores generated by a single classifier did not aid the cause. This is due to the anomalies in results that occur in classification output, which justifies the use of more than one classifiers to obtain the feedback scores.

\begin{figure}[ht!]
    \centering
    \includegraphics[width=0.45\textwidth]{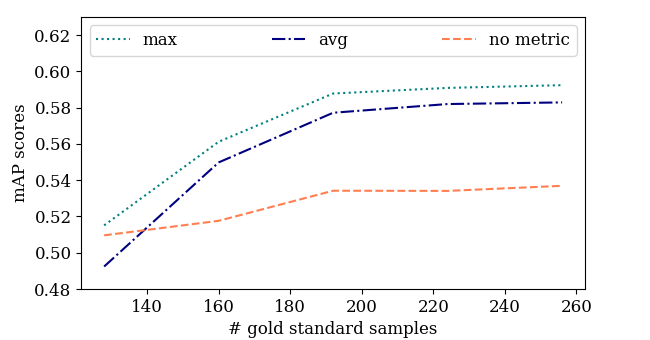}
    \caption{mAP scores compared to size of dataset $\mathcal{S}_1.$}
    \label{fig3}
\end{figure}

Extending the study, experiments were performed by varying the size of the gold standard dataset, $\mathcal{S}_1$, on which $\mathcal{D}_1$ was trained. This experiment shows the factor of performance improvement with an increase in the size of the gold standard dataset. The dataset size was increased from 128 samples up to 256 samples. The scores are reported with the graph in Figure \ref{fig3}. Three scores are reported with `avg' and `max' used as $\alpha(\bar{o}_i)$ compared to the case when $\alpha(\bar{o}_i) = 1$ (no metric). The slope of the graph decreases after increasing $\mathcal{S}_1$ to 192 samples since the size of the generated dataset does not increase much. Below 128 samples, the number of teacher annotations generated was less.



\begin{figure}[ht!]
    \centering
    \includegraphics[width=0.45\textwidth]{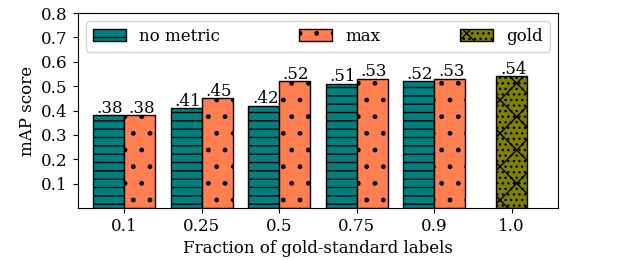}
    \caption{mAP scores on varying fraction of gold standard labels}
    \label{fig4}
\end{figure}

Figure \ref{fig3} compares the mAP scores obtained on varying the fraction of gold standard data to the generated labels. There is a gain of 10\% with half gold standard labels. The scores gradually equate at the extremes and compares the gold standard score towards the end.

%% file: sections/conclusion.tex
\section{Conclusion}
\label{sec7}

The study aimed to obtain good scores with a large number of unlabeled data, using semi-supervision from a teacher, removing the need for manual effort. The human involvement was only for obtaining the gold-standard annotations. Intuitively, the approach discounts the detection algorithm from taking large gradient steps toward a direction that may not be profitable. Better results may be obtained if the gold standard and silver standard data are combined to train the algorithm in the second phase, as is generally done. Here, the scores are reported for silver-standard and gold-standard data separately which gives a better analogy of the performance improvement. It is to be noted that the study can be extended to most detection algorithms; we generically described the loss function in Section \ref{sec4} for the same. If a specific algorithm is used, such as R-CNN, the modifications described in previous texts \cite{sohn2020simple, yang2021interactive} may boost the scores further. Pre-trained CNNs were used for the study, which may not always be the case. For those situations, transfer learning may help, as shown here with the VOC2012 dataset. Although this is not a very good indication of the quality of the label, since there may be anomalies in the classifiers' output as seen in our results, it automates the process and mostly generates good scores. We tried to use more than one classifier to remove the errors that may be generated using a single classifier.